\documentclass[runningheads]{llncs}
\usepackage{graphicx}
\usepackage{amsmath,amssymb,amsfonts}

\begin{document}
\title{Ensemble Forecasting of Monthly Electricity Demand using Pattern Similarity-based Methods
}
\titlerunning{Ensembles Forecasting of MED using PSFMs}
	%
\author{Pawe\l{ }Pe\l ka\and Grzegorz Dudek }

	%
\authorrunning{P. Pe\l ka and G. Dudek}
%
\institute{Electrical Engineering Faculty, Częstochowa University of Technology, \\ Częstochowa, Poland\\
\email{\{p.pelka, dudek\}@el.pcz.czest.pl}}
%
\maketitle              
\begin{abstract}
This work presents ensemble forecasting of monthly electricity demand using pattern similarity-based forecasting methods (PSFMs). PSFMs applied in this study include $k$-nearest neighbor model, fuzzy neighborhood model, kernel regression model, and general regression neural network. An integral part of PSFMs is a time series representation using patterns of time series sequences. Pattern representation ensures the input and output data unification through filtering a trend and equalizing variance. Two types of ensembles are created: heterogeneous and homogeneous. The former consists of different type base models, while the latter consists of a single-type base model. Five strategies are used for controlling a diversity of members in a homogeneous approach. The diversity is generated using different subsets of training data, different subsets of features, randomly disrupted input and output variables, and randomly disrupted model parameters. An empirical illustration applies the ensemble models as well as individual PSFMs for comparison to the monthly electricity demand forecasting for 35 European countries.

\end{abstract}
\section{Introduction}

Load or electricity demand forecasting is an essential tool for power system operation and planning. Mid-term electrical load forecasting (MTLF) involves forecasting the daily peak load for future months as well as monthly electricity demand. MTLF is necessary for maintenance scheduling, fuel reserve planning, hydro-thermal coordination,  electrical energy import/export planning, and security assessment. Deregulated power systems need MTLF to be able to negotiate forward contracts. Therefore, the forecast accuracy translates directly into financial performance for energy market participants.

Methods of MTLF can be divided into a conditional modeling approach and an autonomous modeling approach \cite{Ghi06}. The former focuses on economic analysis and long-term planning of energy policy and uses input variables describing socio-economic conditions, population migrations, and power system and network infrastructure. The latter uses input variables including only historical loads or, additionally, weather factors \cite{Pei11}, \cite{Pel19b}.

For MTLF the classical statistical/econometrics tools are used as well as machine learning tools \cite{Sug11}. The former include ARIMA, exponential smoothing (EST) and linear regression \cite{Bar01}. Problems with adaptability and nonlinear modeling of the statistical methods have increased researchers' interest in machine learning and AI tools \cite{Gon08}. The most popular representatives of these tools are neural networks (NNs) which have very attractive features such as learning capabilities, universal approximation property, nonlinear modeling, and massive parallelism \cite{Chen17}. Among the other machine learning models for MTLF, the following can be mentioned: long short-term memory \cite{Bed18}, weighted evolving fuzzy NNs \cite{Pei11}, support vector machine \cite{Zhao12}, and pattern similarity-based models \cite{Dud15}. 

In recent years ensemble learning has been widely used in machine learning. Ensemble learning systems are composed of many base models. Each of them provides an estimate of a target function. These estimates are combined in
some fashion to produce a common response, hopefully improving accuracy and stability compared to a single learner. The base models can be of the same type (single-model or homogeneous ensemble) or of different types (multi-model or heterogeneous ensemble). The key issue in ensemble learning is ensuring diversity of learners \cite{Bro05}. A good tradeoff between performance and diversity underlies the success of ensemble learning. The source of diversity in the heterogeneous case is a different nature of the base learners. Some experimental results show that heterogeneous ensembles can improve accuracy compared to homogenous ones \cite{Wic03}. This is because the error terms of models of different types are less correlated than the errors of models of the same type. 
Generating diverse learners which give uncorrelated errors in a homogeneous ensemble is a challenging problem. Diversity can be achieved through several strategies. One of the most popular is learning on different subsets of the training
set or different subsets of features. Other common approaches include using different values of hyperparameters and parameters of learners. In the field of forecasting, it was shown that ensembling of the forecasts enhances the robustness of the model, mitigating the model and parameter uncertainty \cite{Pet18}.  

In this work we build heterogeneous and homogeneous ensembles for MTLF using pattern similarity-based forecasting models (PSFMs) \cite{Dud15} as base learners. PBSMs turned out to be very effective models (accurate and simple) for both mid and short-term load forecasting \cite{Dud17}, \cite{Dud15b}. In this study, we investigate what profit we will achieve from an ensembling of the forecasts generated by PBFMs. For heterogeneous ensemble, we employ $k$-nearest neighbor model, fuzzy neighborhood model, kernel regression model, and general regression neural network. For homogeneous ensemble, we employ a fuzzy neighborhood model and generate its diversity using five strategies.     

The remainder of this paper is structured as follows. In Section 2, we present pattern representation of time series, a framework of the pattern similarity-based forecasting, and PSFMs. In Section 3, we describe heterogeneous and homogeneous ensemble forecasting using PSFMs. Section 4 shows the setup of the empirical experiments and the results. Finally, Section 5 presents our conclusion. 

\section{Pattern Similarity-based Forecasting}

\subsection{Pattern Representation of Time Series}

Monthly electricity demand time series express a trend, yearly cycles, and random component. To deal with seasonal cycles and trends in our earlier work, we proposed similarity-based models operating on patterns of the time series sequences \cite{Dud15}, \cite{Dud17}. The patterns filter out the trend and those seasonal cycles longer than the basic one and even out variance. They also ensure the unification of input and output variables. Consequently, pattern representation simplifies the forecasting problem and allows us to use models based on pattern similarity.

Input pattern $\mathbf{x}_i = [x_{i,1} x_{i,2} … x_{i,n}]^T$ is a vector of predictors representing a sequence $ X_i = \{E_{i–n+1}, E_{i–n+2},…, E_i \} $ of $n$ successive time series elements $E_i$ (monthly electricity demands) preceding a forecasted period. In this study we use the following definition of x-pattern components:

\begin{equation}\label{eq4}
x_{i,t} = \frac{E_{i-n+t}-\overline{E}_i}{D_i}
\end{equation}
where $t = 1, 2, ..., n$, $\overline{E}_i$ is a mean of sequence $X_i$, and $D_i = \sqrt{\sum_{j=1}^{n} (E_{i-n+j}-\overline{E}_i)^2}$ is a measure of its dispersion.

The x-pattern defined using \eqref{eq4} is a normalized vector composed of the elements of sequence $X_i$. Note that the original time series sequences $X_i$ having different mean and dispersion are unified, i.e. they are represented by x-patterns which all have zero mean, the same variance and also unity length. 

Output pattern $\mathbf{y}_i = [y_{i,1} y_{i,2} … y_{i,m}]^T$ represents a forecasted sequence of length $m=12$: $ Y_i = \{ E_{i+1}, E_{i+2},…, E_{i+m}\} $. The output pattern is defined similarly to the input one: 

\begin{equation}\label{eq8}
y_{i,t} = \frac{E_{i+t}-\overline{E}_i^*}{D_i^*}
\end{equation}
where $t = 1, 2, ..., m$, and $\overline{E}_i^*$ and $D_i^*$ are coding variables described below.

 Two variants of the output patterns are considered. In the first one, denoted as V1, the coding variables, $\overline{E}_i^*$ and $D_i^*$, are the mean and dispersion, respectively, of the forecasted sequence $Y_i$. But in this case, when the forecasted sequence $Y_i$ of the monthly electricity demands is calculated from the forecasted y-pattern, $\mathbf{\widehat{y}_i}$, using transformed equations \eqref{eq8}:     

\begin{equation}\label{eq9}
\widehat{E}_{i+t} = \widehat{y}_{i,t}D_i^* + \overline{E}_i^*, \quad t = 1, 2, ..., m
\end{equation}
the coding variables are not known, because they are the mean and dispersion of future sequence $Y_i$, which has just been forecasted. In this case, the coding variables are predicted from their historical values. In the experimental part of the work, the coding variables are predicted using ARIMA and ETS.

To avoid forecasting the coding variables we use another approach. Instead of using the mean and dispersion of the forecasted sequence $Y_i$ as coding variables, we introduce in \eqref{eq8} and \eqref{eq9} as coding variables the mean and dispersion of sequence $X_i$, i.e. $\overline{E}_i^*=\overline{E}_i$, $D_i^*=D_i$. When the PSFM generates the forecasted y-pattern, the forecast of the monthly demands are calculated from \eqref{eq9} using known coding variables for the historical sequence $X_i$. This variant of the y-pattern definition is denoted as V2. 

\subsection{Forecasting Models}

Pattern similarity-based forecasting procedure can be summarized in the following steps \cite{Dud15}:

\begin{enumerate}
	\item Mapping the original time series sequences into x- and y-patterns.
	\item Selection of the training x-patterns similar to the query pattern $\mathbf{x}$.
	\item Aggregation of the y-patterns paired with the similar x-patterns to obtain the forecasted pattern $\widehat{\mathbf{y}}$.
	\item Decoding pattern $\widehat{\mathbf{y}}$ to get the forecasted time series sequence $\widehat{Y}$.
\end{enumerate} 

In step 3, y-patterns are aggregated using weights which are dependent on the similarity between a query pattern $\mathbf{x}$ and the training x-patterns. The regression model mapping x-patterns into y-patterns is of the form:

\begin{equation}\label{eq10}
m(\mathbf{x})= \sum\limits_{i=1}^N w(\mathbf{x}, \mathbf{x}_i)\mathbf{y}_i
\end{equation} 
where $\sum_{i=1}^N w(\mathbf{x}, \mathbf{x}_i)=1$, $w(.,.)$ is a weighting function.

Model \eqref{eq10} is nonlinear if $w(.,.)$ maps $\mathbf{x}$ nonlinearly. Different definitions of $w(.,.)$ are presented below where the PSFMs are specified.

\subsubsection{$k$-Nearest Neighbor Model} 

estimates $m(.)$ as the weighted average of the y-patterns in a varying neighborhood of query pattern $\mathbf{x}$ (this model is denoted as $k$-NNw). The neighborhood is defined as a set of $k$ nearest neighbors of $\mathbf{x}$ in the training set $\Phi$. The regression function is as follows:

\begin{equation}\label{eq11}
m(\mathbf{x})= \sum\limits_{i \in \Omega_k(\mathbf{x})} w(\mathbf{x}, \mathbf{x}_i) \mathbf{y}_i
\end{equation} 
where $\Omega_k(\mathbf{x})$ is a set of indices of $k$ nearest neighbors of $\mathbf{x}$ in $\Phi$ and the weighting function is of the form \cite{Dud15b}:
\begin{equation}\label{eq12}
w(\mathbf{x}, \mathbf{x}_i)=\frac{v(\mathbf{x}, \mathbf{x}_i)}{\sum\limits_{j \in \Omega_k(\mathbf{x})}v(\mathbf{x}, \mathbf{x}_j)}
\end{equation}

\begin{equation}\label{eq13}
v(\mathbf{x}, \mathbf{x}_i)=\rho \left( \frac{1-d(\mathbf{x},\mathbf{x}_i)/d(\mathbf{x},\mathbf{x}^k)}{1+\gamma d(\mathbf{x},\mathbf{x}_i)/d(\mathbf{x},\mathbf{x}^k)} -1\right) +1
\end{equation}
where $\mathbf{x}^k$  is the $k$-th nearest neighbor of $\mathbf{x}$ in $\Phi$, $d(\mathbf{x},\mathbf{x}_i)$ is a Euclidean distance between $\mathbf{x}$ and its $i$-th nearest neighbor, $\rho \in [0, 1]$ is a parameter deciding about the differentiation of weights, and $\gamma \geq -1$ is a parameter deciding about a convexity of the weighting function.

\subsubsection{Fuzzy Neighborhood Model}

(FNM) takes into account all training patterns when constructing the regression surface \cite{Pel17}. In this case, all training patterns belong to the query pattern neighborhood, with a different membership degree. The membership function is dependent on the distance between the query pattern $\mathbf{x}$ and the training pattern $\mathbf{x}_i$ as follows:

\begin{equation}\label{eq14}
\mu (\mathbf{x},\mathbf{x}_i)=\exp \left(-\left( \frac{d(\mathbf{x},\mathbf{x}_i)}{\sigma}
\right)^\alpha \right) 
\end{equation}
where $\sigma$ and $\alpha$ are parameters deciding about the membership function shape. 

The weighting function in FNM is as follows:

\begin{equation}\label{eq15}
w(\mathbf{x}, \mathbf{x}_i)=\frac{\mu (\mathbf{x},\mathbf{x}_i)}{\sum\limits_{j = 1}^N  \mu(\mathbf{x},\mathbf{x}_j)}
\end{equation}

Membership function \eqref{eq14} is a Gaussian-type function. The model parameters, $\sigma$ and $\alpha$, shape the membership function and thus control the properties of the estimator.

\subsubsection{Nadaraya-Watson Estimator}

(N-WE) estimates regression function $m(.)$ as a locally weighted average, using in \eqref{eq10} a kernel $K_h$ as a weighting function: 

\begin{equation}\label{eq16}
w(\mathbf{x}, \mathbf{x}_i)=\frac{K_h (\mathbf{x}-\mathbf{x}_i)}{\sum\limits_{j = 1}^N  K_h(\mathbf{x}-\mathbf{x}_j)}
\end{equation}

When the input variable is multidimensional, the kernel has a product form. In such a case, for a normal kernel, which is often used in practice, the weighting function is defined as \cite{Dud15b}, \cite{Dud17}:

\begin{equation}\label{eq17}
w(\mathbf{x}, \mathbf{x}_i)=\frac{\exp \left(-\sum\limits_{t = 1}^n 
	\frac{(x_t-x_{i,t})^2}{2h_t^2} \right)}
{\sum\limits_{j = 1}^N \exp \left(-\sum\limits_{t = 1}^n 
	\frac{(x_t-x_{j,t})^2}{2h_t^2} \right)} 
\end{equation}
where $h_t$ is a bandwidth for the $t$-th dimension.

The bandwidths decide about the bias-variance tradeoff of the estimator. 

\subsubsection{General Regression Neural Network Model} (GRNN) is composed of four layers: input, pattern (radial basis layer), summation and output layer \cite{Pel19}. The pattern layer transforms inputs nonlinearly using Gaussian activation functions of the form:

\begin{equation}\label{eq18}
G(\mathbf{x},\mathbf{x}_i)=\exp \left(-\frac{\| \mathbf{x}-\mathbf{x}_i) \| ^2}{\sigma_i^2} \right)
\end{equation}
where $\|.\|$ is a Euclidean norm and $\sigma_i$ is a bandwidth for the $i$-th pattern.

The Gaussian functions are centered at different training patterns $\mathbf{x}_i$. The neuron output expresses a similarity between the query pattern and the $i$-th training pattern. This output is treated as the weight of the $i$-th y-pattern. So the pattern layer maps the $n$-dimensional input space into $N$-dimensional space of similarity, where $N$ is a number of training patterns. The weighting function implemented in GRNN is defined as:

\begin{equation}\label{eq19}
w(\mathbf{x}, \mathbf{x}_i)=\frac{G(\mathbf{x},\mathbf{x}_i)}{\sum\limits_{j = 1}^N  G(\mathbf{x},\mathbf{x}_j)}
\end{equation}

The performance of PSFMs is related to the weighting function parameters governing the smoothness of the regression function \eqref{eq10}. For wider weighting function the model tends to increase bias and decrease variance. Thus, too wide weighting function leads to oversmoothing, while too narrow weighting function leads to undersmoothing. The PSFM parameters should be adjusted to the target function.

\section{Ensemble Forecasting using PSFMs}
Two approaches for ensemble forecasting are used: heterogeneous and homogeneous. The former consists of different base models, while the latter consists of a single-type base model. In the heterogeneous approach, we use the PSFMs described above as base models. A diversity of learners, which is the key property that governs an ensemble performance, in this case, results from different types of learners. 

To control the diversity in the homogeneous approach we use the following strategies \cite{Dud17b}:   

\begin{enumerate}
	\item Learning on different subsets of the training data. For each ensemble member a random training sample without replacement of size $N' < N$ is selected from the training set $\Phi$. 
	\item Learning on different subsets of features. For each ensemble member the features are randomly sampled without replacement. The sample size is $n' < n$. In this case, the optimal model parameters may need correction for ensemble members due to a reduction in Euclidean distance between x-patterns in $n'$-dimensional space relative to $n$-dimensional space. 
	\item Random disturbance of the model parameters. For FNM the initial value of width $\sigma$ is randomly perturbed for $k$-th member by a Gaussian noise: $\sigma_k= \sigma \cdot \xi_k$, where $\xi_k \sim N(0,\sigma_s)$. 
	\item Random disturbance of x-patterns. The components of x-patterns are perturbed for $k$-th member by a Gaussian noise: $x_{i,t}^k = x_{i,t} \cdot \xi_{i,t}^k$, where $\xi_{i,t}^k \sim N(0,\sigma_x)$. 
	\item Random disturbance of y-patterns. The components of y-patterns are perturbed for $k$-th member by a Gaussian noise: $y_{i,t}^k = y_{i,t} \cdot \xi_{i,t}^k$, where $\xi_{i,t}^k \sim N(0,\sigma_y)$. 
\end{enumerate}

Standard deviations of the noise signals, $\sigma_s, \sigma_x, \sigma_y$, control the noise level and are selected for each forecasting task as well as $N'$ and $n'$.

The first strategy controlling diversity is similar to bagging \cite{Bre96}, where the predictors are built on bootstrapped versions of the original data. In bagging, unlike our approach, the sample size is $N' = N$ and the random sample is drawn with replacement. The second strategy is inspired by the random subspace method \cite{Ho98} which is successfully used to construct random forests, very effective tree-based classification and regression models. Note that the diversity of learners has various sources. They include data uncertainty (learning on different subsets of the training set, learning on different features of x-patterns, learning on disturbed input and output variables) and parameter uncertainty.    
 
The forecasts of y-patterns generated by $K$ base models, $\mathbf{\widehat{y}}_k$, are aggregated using simple averaging to obtain an ensemble forecast:

\begin{equation}\label{eqA}
\mathbf{\widehat{y}} = \frac{1}{K}\sum_{k=1}^{K} \mathbf{\widehat{y}}_k
\end{equation} 

In this study we use the mean for aggregation, but also other functions, such as median, mode, or trimmed mean could be used. As shown in \cite{Cha18} a simple average of forecasts often outperforms forecasts from single models and a more complicated weighting scheme does not always perform better than a simple average. 

When the y-pattern is determined from \eqref{eqA}, the forecasted monthly demands $\mathbf{\widehat{E}}$ are calculated from \eqref{eq9} using coding variables which are determined from history or predicted, depending on the model variant, V1 or V2. 

\section{Simulation Study}

In this section, we apply the proposed ensemble forecasting models to mid-term load forecasting using real-world data: monthly electricity demand time series for 35 European countries. The data are taken from the publicly available ENTSO-E repository (www.entsoe.eu). The time series differ in levels, trends, variations and yearly shapes. They differ also in a length, i.e. they cover: 24 years for 11 countries, 17 years for 6 countries, 12 years for 4 countries, 8 years for 2 countries, and 5 years for 12 countries. The models forecast for the twelve months of 2014 (last year of data) using data from the previous period for training.

We built four heterogeneous ensembles:
\begin{description}
	\item [Ensemble1] composed of PSFMs described in Section 3, i.e. $k$-NNw, FNM, N-WE and GRNN, which are trained on the y-patterns defined with the coding variables determined from the historical sequence $X_i$ (y-pattern definition V2).
	\item [Ensemble2] composed of PSFMs which are trained on the y-patterns defined with the coding variables predicted for the forecasted sequence $Y_i$ using ARIMA (y-pattern definition V1). The base models, in this case, are denoted as $k$-NNw+ARIMA, FNM+ARIMA, N-WE+ARIMA and GRNN+ARIMA. 
	\item [Ensemble3] composed of PSFMs which are trained on the y-patterns defined in the same way as for Ensemble2, but the coding variables are predicted using ETS. The base models, in this case, are denoted as $k$-NNw+ETS, FNM+ETS, N-WE+ETS and GRNN+ETS.
	\item [Ensemble4] composed of all variants of PSFM models mentioned above for Ensemble1, Ensemble2 and Ensemble3, i.e. twelve models.
\end{description}

For prediction the coding variables we used the ARIMA and ETS implementations in R statistical software environment: functions \texttt{auto.arima} and \texttt{ets} from the \texttt{forecast} package. These functions implement automatic ARIMA and ETS modeling, respectively, and identify optimal models estimating their parameters using Akaike information criterion (AICc) \cite{Hyn19}.

The optimal values of hyperparameters for each PSFM were selected individually for each of 35 time series in the grid search procedure using cross-validation. These hyperparameters include: length of the x-patterns $n$, number of nearest neighbors $k$ in $k$-NNw (linear weighting function was assumed with $\rho=1$ and $\gamma=0$), width parameter $\sigma$ in FNM (we assumed $\alpha=2$), bandwidth parameters $h_t$ in N-WE, and bandwidth $\sigma$ in GRNN.

\begin{figure}[htbp]
	\centerline{\includegraphics[width=1\textwidth]{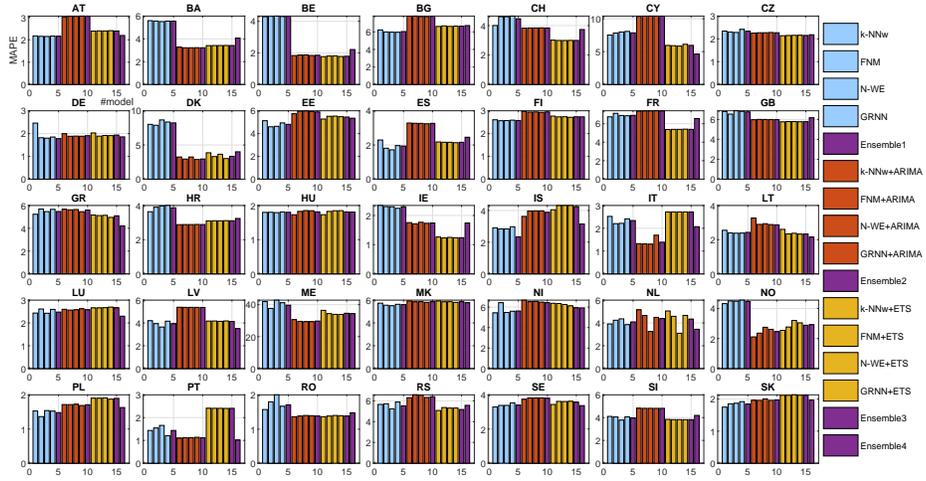}}
	\caption{MAPE for each country.}
	\label{figCount}
\end{figure}

The forecasting errors on the test sets (mean absolute percentage error, MAPE) for each model and each country are shown in Fig. \ref{figCount} and their averaged values are shown in Table 1. In Table 1 also median of the absolute percentage error (APE), interquartile ranges of APE and root mean square error (RMSE) averaged over all countries are shown.
The forecasting accuracy depends heavily on the variant of the coding variables determination. The most accurate variant on average is V1+ETS and the least accurate is variant V2 where coding variables are determined from history. 

\begin{figure}[htbp]
	\centerline{\includegraphics[width=0.69\textwidth]{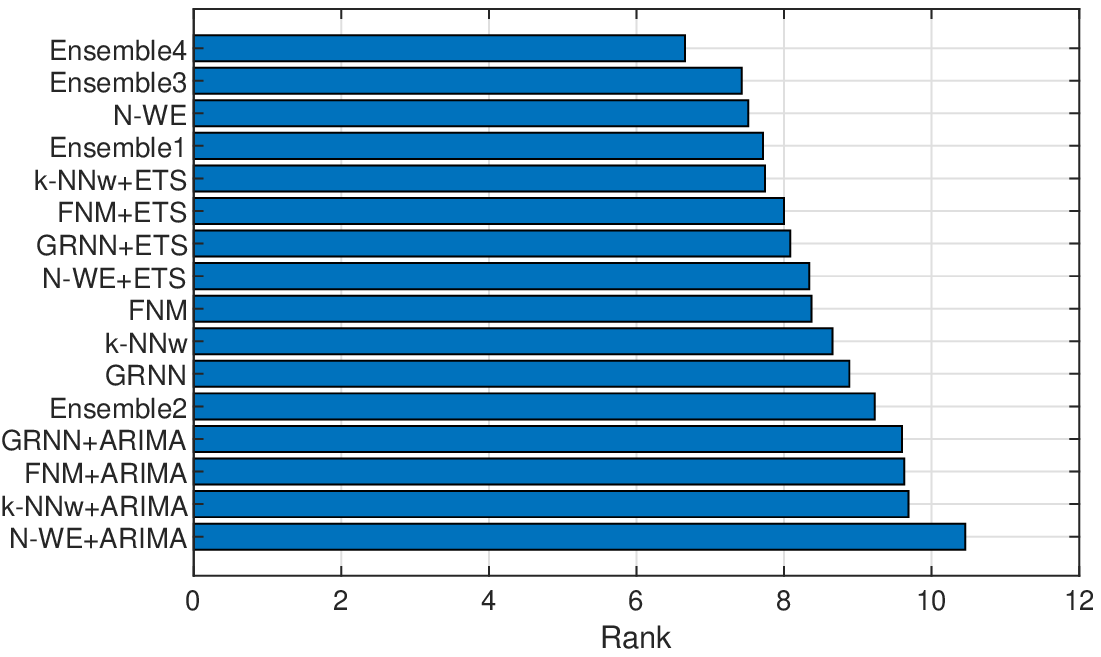}}
	\caption{Ranking of the models.}
	\label{figRank}
\end{figure}

Fig. \ref{figRank} shows the ranking of the models based on MAPE. The rank is calculated as the average rank of the model in the rankings performed individually for each country. As you can see from this figure, the Ensemble4 and Ensemble3 models were the most accurate for the largest number of countries. Model N-WE took the third position. Note that ensemble combining the group of PSFMs (V1+ARIMA, V1+ETS or V2) occupies a higher position in the ranking than individual members of this group. The exception is N-WE which achieves better results than Ensemble1. A similar conclusion can be drawn from the ranking based on RMSE.

\begin{table}[]
	\caption{Results for base models and heterogeneous ensembles.}
	\begin{center}
	\begin{tabular}{|l|c|c|c|c|}
		\hline
		Model       & Median \textit{APE} & { }{ }\textit{MAPE}{ }{ } & { }{  }\textit{IQR}{ }{ } &{ }{ }\textit{RMSE}{ }{ }\\ \hline
		k-NNw       & 2.89       & 4.99 & 4.06 & 368.79 \\
		FNM         & 2.88       & 4.88 & 4.43 & 354.33 \\
		N-WE        & 2.84       & 5.00 & 4.14 & 352.01 \\
		GRNN        & 2.87       & 5.01 & 4.30 & 350.61 \\
		Ensemble1   & 2.88       & 4.90 & 4.13 & 351.89 \\
		k-NNw+ARIMA & 2.89       & 4.65 & 4.02 & 346.58 \\
		FNM+ARIMA   & 2.87       & 4.61 & 3.83 & 341.41 \\
		N-WE+ARIMA  & 2.85       & 4.59 & 3.74 & 340.26 \\
		GRNN+ARIMA  & 2.81       & 4.60 & 3.77 & 345.46 \\
		Ensemble2   & 2.90       & 4.60 & 3.84 & 342.43 \\
		k-NNw+ETS   & 2.71       & 4.47 & 3.43 & 327.94 \\
		FNM+ETS     & 2.64       & 4.40 & 3.34 & 321.98 \\
		N-WE+ETS    & 2.68       & 4.37 & 3.20 & 320.51 \\
		GRNN+ETS    & 2.64       & 4.38 & 3.35 & 324.91 \\
		Ensemble3   & 2.64       & 4.38 & 3.40 & 322.80 \\
		Ensemble4   & 2.70       & 4.31 & 3.49 & 327.61 \\ \hline
	\end{tabular}
	\label{tab1}
	\end{center}
\end{table}

The homogeneous ensembles were built using FNM in variant V2 as a base model. Five strategies of diversity generation described in Section 3 were applied. Ensembles constructed in this way are denoted as FNMe1, FNMe2, ...FNMe5. In the FNMe2 case, where the diversity is obtained by selection $n'$ x-pattern components, the optimal width parameter $\sigma$ (selected for a single FNM) is corrected for ensemble members by the factor $(n'/n)^{0.5}$. This is due to a reduction in Euclidean distance between x-patterns in $n'$-dimensional space relative to the original $n$-dimensional space. 

The forecasts were generated independently by each of $K=100$ ensemble members. Then the forecasts were combined using \eqref{eqA}. The following parameters of the ensembles were selected on the training set using a grid search: 
\begin{itemize}
	\item size of the random sample of training patterns in FNMe1: $N'=0.85N$,
	\item size of the random sample of features in FNMe2: $n'=0.925n$,
	\item standard deviation of the disruption of width parameter $\sigma$ in FNMe3: $\sigma_s=0.475$,
	\item standard deviation of the disruption of x-patterns in FNMe4: $\sigma_x=0.4$,
	\item standard deviation of the disruption of y-patterns in FNMe5: $\sigma_y=0.65$.
\end{itemize}

Table 2 shows the results for FNM ensembles. It can be seen from this table that the errors for different sources of diversity are similar. It is hard to indicate the best strategy for member diversification. When comparing results for FNM ensembles and single base model FNM (see Table 1), we can see a slightly lower MAPE for ensembles with the exception of FNMe5 where MAPE is the same as for FNM. But RMSE is higher for ensemble versions of FNM than for single FNM. 

\begin{figure}[htbp]
	\centerline{\includegraphics[width=0.69\textwidth]{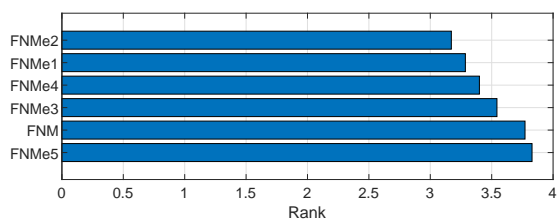}}
	\caption{Ranking of the FNM homogeneous ensembles.}
	\label{figRank2}
\end{figure}

Fig. \ref{figRank2} shows the ranking of the FNM ensembles based on MAPE. The ensembles FNMe1-FNMe4 are more accurate then FNM for most countries. Ensemble FNMe5 turned out to be less accurate than FNM. A similar conclusion can be drawn from the ranking based on RMSE.      

\begin{table}[]
	\caption{Results for FNM homogeneous ensembles.}
	\begin{center}
	\begin{tabular}{|l|c|c|c|c|}
		\hline
		Model & Median \textit{APE} & { }{ }\textit{MAPE}{ }{ } & { }{  }\textit{IQR}{ }{ } &{ }{ }\textit{RMSE}{ }{ }\\ \hline
		FNMe1 & 2.88       & 4.84 & 4.18 & 370.52 \\
		FNMe2 & 2.85       & 4.84 & 4.06 & 366.35 \\
		FNMe3 & 2.80       & 4.83 & 4.23 & 371.94 \\
		FNMe4 & 2.90       & 4.86 & 4.10 & 373.73 \\
		FNMe5 & 2.97       & 4.88 & 4.18 & 375.84 \\ \hline
	\end{tabular}
	\label{tab2}
	\end{center}
\end{table}

\section{Conclusion}
Ensemble forecasting is widely used for improving the forecast accuracy over the individual models. In this work, we investigate single-model and multi-model ensembles based on pattern-similarity forecasting models for mid-term electricity demand forecasting. The key issue in ensemble learning is ensuring the diversity of learners. The advantage of heterogeneous ensembles is that the errors of the base models are to be weakly correlated because of the different nature of the models. But in our case the PSFMs are similar in nature, so we can expect error correlation. The results of simulations do not show a spectacular improvement in accuracy for homogeneous ensemble comparing to its members. However, the ranking shown in Fig. \ref{figRank} generally confirms better results for ensembles than for their members. 

In homogeneous ensembles, we can control a diversity level of members. We propose five strategies for this including strategies manipulating training data and model parameters. Among them, strategies based on learning on different subsets of training data and different subsets of features turned out to be most effective.

\end{document}